# Toward Full Autonomous Laboratory Instrumentation Control with Large Language Models

Yong Xie[1,2], Kexin He[1], Andres Castellanos-Gomez[2]

[1] Key Laboratory of Wide Band-Gap Semiconductor Technology, School of Advanced Materials and Nanotechnology, Xidian University, Xi'an 710071, China;

[2] 2D Foundry Research Group. Instituto de Ciencia de Materiales de Madrid (ICMM-CSIC), Madrid, E-28049, Spain.

yxie@xidian.edu.cn  andres.castellanos@csic.es

## ABSTRACT

The control of complex laboratory instrumentation often requires significant programming expertise, creating a barrier for researchers lacking computational skills. This work explores the potential of large language models (LLMs), such as ChatGPT, to enable efficient programming and automation of scientific equipment. Through a case study involving the implementation of a setup that can be used as a single-pixel camera or a scanning photocurrent microscope, we demonstrate how ChatGPT can facilitate the creation of custom scripts for instrumentation control, significantly reducing the technical barrier for experimental customization. Building on this capability, we further illustrate how LLM-assisted tools can be used to develop autonomous agents capable of independently operating laboratory instruments. This approach underscores the transformative role of LLM-based tools in democratizing laboratory automation and accelerating scientific progress.

**Introduction**

Scientific instrumentation is a cornerstone of modern research, enabling discoveries across a wide range of disciplines. Innovations such as the scanning tunneling microscope (STM), super-resolution microscopy, and transmission electron microscopy (TEM) have enabled atomic-scale imaging, advanced biological imaging, and structural analysis of quantum materials, respectively [1-9]. These breakthroughs exemplify how instrumentation drives progress by providing researchers with tools to probe the unknown.

However, while the development of new instruments is critical, effectively utilizing existing laboratory equipment remains challenging for many research groups. Scientific instruments often rely on sophisticated control software for workflow automation and data acquisition [10]. Developing such software traditionally demands significant programming expertise, creating a barrier for researchers who may lack computational backgrounds. Consequently, many laboratories depend on commercially available equipment with proprietary software. While these tools simplify basic operation, they often lack the flexibility needed for custom experimental setups, limiting their potential for innovation.

Recent advancements in artificial intelligence (AI) have revolutionized scientific workflows, enabling tasks such as predicting material properties, uncovering structure-property relationships, and accelerating materials discovery [11-13]. However, these AI-driven solutions frequently require custom software and programming expertise, rendering them inaccessible to groups without strong computational resources. This gap highlights the need for more accessible approaches to leveraging AI technologies for laboratory automation and instrumentation control.

Large language model (LLM)-based tools, such as ChatGPT, provide a promising solution to this challenge. These models have already demonstrated their utility in diverse research tasks, including drafting scientific manuscripts [14,15], uncovering structure-property relationships [16], proposing novel scientific hypotheses [17,18], and even contributing as peer reviewers [19, 20]. Beyond these applications, LLM-based tools offer an intuitive interface for generating and refining programming code, streamlining the automation of experimental workflows. By translating natural language instructions into executable commands, these models empower researchers to communicate effectively with scientific equipment, enabling custom workflows without extensive programming knowledge. For instance, systems like ORGANA (a robotic assistant that autonomously performs lab tasks based on natural language instructions) demonstrate how LLM-based tools can automate diverse chemistry experiments, reducing manual workload and enhancing efficiency [21]. Similarly, studies have highlighted the role of LLM-based tools in accelerating materials discovery through high-throughput experimentation and data-driven strategies. These developments signify a shift towards more autonomous, flexible, and efficient laboratory environments, aligning with the goals of self-driving labs and AI-assisted research [21-24].

LLM-based tools can also address a critical pain point in laboratories: the difficulty of controlling specialized instruments that are poorly supported by user communities. To date, LLM-based programming has already demonstrated control capabilities over relatively simple devices (e.g. power supplies, robotic arms, or basic measurement units) typically through static scripts with limited interaction or adaptability [25-27]. More significantly, only a few studies have systematically explored the use of AI agents capable of autonomously interpreting user intent, managing experimental context, and executing complete control sequences, representing an essential step toward achieving intelligent and autonomous laboratory systems.

In this paper, we demonstrate the application of ChatGPT to streamline the control of commercially available scientific equipment. Through a case study involving the control of a set of hardware that can be used to implement a single-pixel camera system or a scanning photocurrent microscope, we illustrate how researchers can use LLM-based tools to iteratively develop functional scripts, simplifying the programming of complex setups. This approach highlights the potential of LLM-based tools to democratize laboratory automation and make advanced experimental configurations accessible to a wider scientific community.

**Results and Discussion**

We begin with the implementation of the hardware setup to demonstrate a single-pixel camera system, providing an illustrative example of how LLM-based tools can assist in hardware control and automation. Figure 1 outlines the pipeline for using LLM-based tools in instrumentation control. The process starts with the researcher selecting a programming language (MATLAB or Python) and planning the hardware configuration. We assembled a minimal working setup to construct either a scanning photocurrent system or a single-pixel camera, both widely used in optoelectronics research [28].

In this setup, the Keithley 2450 source measure unit (SMU) applies a bias voltage and measures the current of a photodetector, while a motorized stage (Standa 8MTF-75LS05, 107291, 107293, Standa Controller model 8SMC4-USB-B9-2) provides precise XY movement. The communication protocols for the equipment were identified based on manufacturer documentation: the Keithley 2450 communicates via VISA protocol over USB connection, and the Standa XY stage communicates through a USB serial (COM) port.

Control scripts for both instruments were iteratively generated through interactions with ChatGPT. While the VISA-based communication of the Keithley SMU is widely supported by general-purpose instrument libraries, the Standa stage required handling a more specialized protocol. To demonstrate the versatility of the LLM-based approach, we applied the same prompt-driven code generation method to this non-standard interface, incorporating information from the manufacturer's documentation into the LLM prompts and refining the code iteratively based on system feedback. Once data were collected, the datasets were saved and visualized using scripts refined through user prompts.

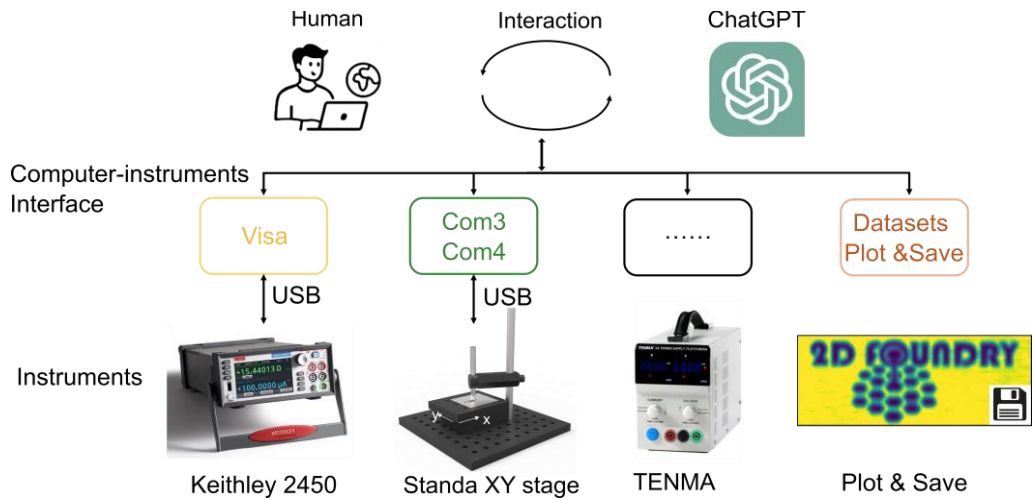

**Figure 1. Workflow of the LLM (ChatGPT) based instrumentation control.** After identifying the computer-instrument interfaces, the instruments are controlled using scripts generated in response to human prompts through a large language model (e.g., ChatGPT).

After assembling the hardware components and making the necessary connections, we outlined the system's operation to guide code generation. The XY stage moves the sample in a raster scan pattern, covering the entire area to be imaged. At each pixel position, the stage halts momentarily for measurement. The SMU applies a bias voltage to the CdS photodetector and measures the photocurrent generated by light reflected from the sample. A fiber-based reflection probe (Thorlabs RP24) connected to an LED is used for illumination and collection of reflected light. This photocurrent directly correlates with the intensity of light reflected from the sample at each pixel. As the scan progresses, variations in photocurrent are recorded, enabling the construction of a detailed reflectance map that highlights areas with differing optical properties and reveals surface features.

Figure 2a shows a flowchart that outlines how the XY stage is integrated with the Keithley 2450 source measurement unit, breaking the instrumentation control into distinct tasks. At each step in the flowchart, a user prompt generated by the LLM helps configure and test the relevant hardware control or data acquisition code. Figure 2b presents a typical example of such a prompt, which focuses on dividing the coding procedure into small, manageable segments. This segmentation enables rapid testing and iterative user feedback at each stage, ensuring that any issues can be identified and resolved before proceeding.

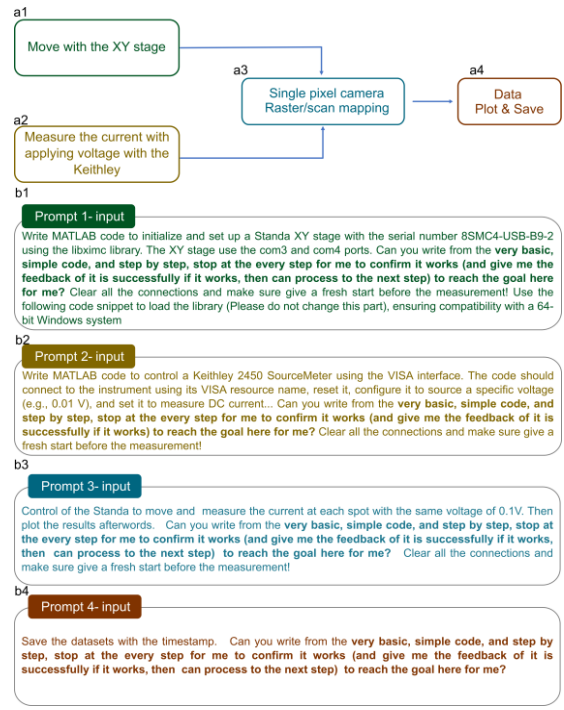

**Figure 2. Typical prompts used in LLM-based instrumentation illustrate the STEP approach (Segment, Test, Evaluate, Proceed).** (a) The flow chart of the process to combine the movement of stage with the Keithley 2450 source measurement unit. (b) The typical prompt used for the corresponding interaction with each block in (a). The key point is breaking the coding process into small steps so the user can quickly review and adjust the code through interaction with ChatGPT.

A key factor in this iterative approach is obtaining feedback early in the process to verify that the generated code functions correctly. To achieve this balance between efficiency and accuracy, the STEP Approach (Segment, Test, Evaluate, Proceed) was developed. The typical prompt "*Can you write from the very basic, simple code, and step by step, stop at every step for me to confirm it works (and give me the feedback if it is successful, then proceed to the next step)?*" was employed during the first three steps shown in Figure 2. This approach demonstrates how LLM-based methods can simplify and streamline instrument control, even for users with limited programming experience.

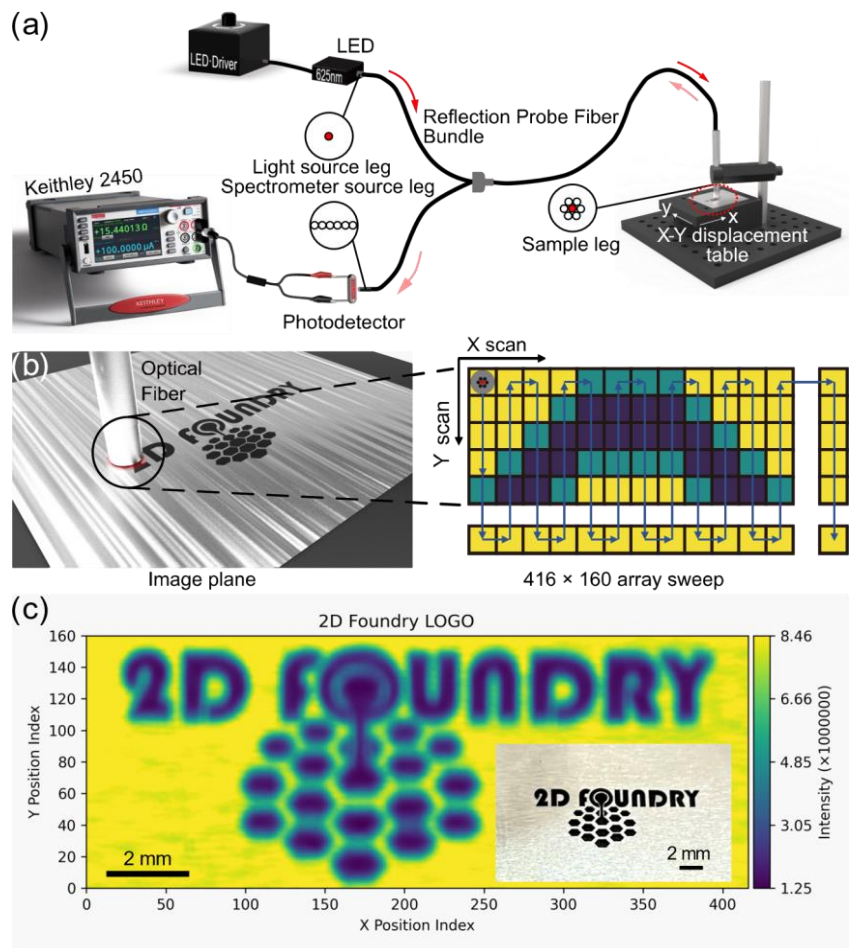

**Figure 3. The hardware and results of the scanning photocurrent image building using the ChatGPT−based instrumentation.** (a) The hardware assembly of the LED, photodetector, and the scanning Standa XY stage connected with the Thorlabs reflection probe with linear leg (RP25). (b) A zoomed-in 3D representation of the optical fiber on the "2D Foundry" logo. A cartoon explaining the snake-like pattern raster scanning is shown besides. (c) The scanning photocurrent image captured by the code generated with ChatGPT. The inset shows a photograph of the test sample, fabricated by laser cut aluminum foil onto a black paper.

Figure 3(a) illustrates the hardware configuration used to implement a single-pixel camera, which includes the fiber coupled LED light source, the Standa XY stage, the photodetector, and the Keithley 2450 source measurement unit. The optical path relies on a Thorlabs reflection probe (RP25) with a leg that directs and collects the light signal. Figure 3(b) provides a 3D rendering, showing the fiber positioned over the sample. The setup employs a "snake-like" raster scanning, as depicted in Figure 3(c). The code generated by ChatGPT acquires and plots the measured photocurrents, producing a final image of the scanned region. The inset of Figure 3(c) shows an actual photograph of the laser-cut aluminum foil mounted on black paper, which corresponds well with the features observed in the scanning photocurrent map. This demonstration highlights how LLM-based instrumentation can streamline optoelectronic measurements and data visualization, even for users with minimal coding experience. Scanning photocurrent mapping systems are widely used as spatially resolved tools for probing photoresponse channels [10]. Figure 4 illustrates the adaptation of the ChatGPT-based instrumentation to perform scanning photocurrent mapping with minor modifications to the setup. In this configuration, a fiber coupled LED light source illuminates with a focused spot, using a focusable collimator, onto the surface of a CdS photodetector. During the raster scanning process, photocurrent measurements are recorded using the Keithley 2450 while the Standa XY stage moves across the sample. The same ChatGPT-generated code was employed for this setup, enabling a "snake-like" raster scanning pattern with $200 \times 120$ steps. The resulting photocurrent mapping, shown in Figure 4(c), clearly resolves spatial inhomogeneities in the photodetector's photocurrent response. This demonstration highlights the effectiveness of ChatGPT-based instrumentation for producing high-resolution, spatially resolved measurements, even in straightforward experimental setups.

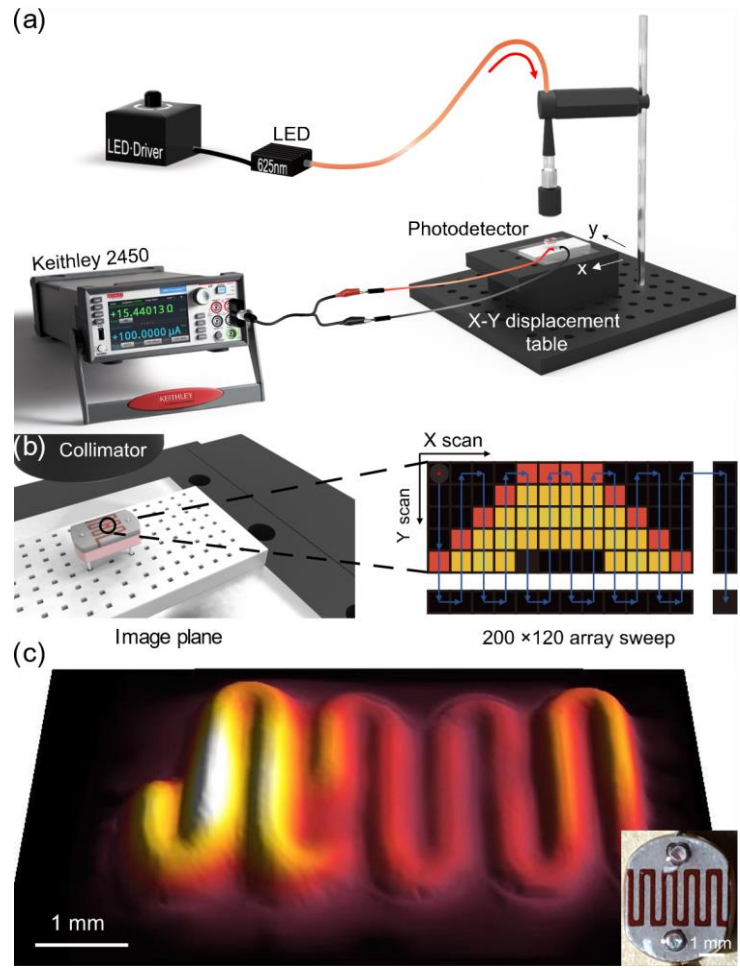

**Figure 4. Scanning photocurrent mapping of a commercial photodetector using the ChatGPT-based instrumentation.** (a) The hardware assembly includes an LED light source, a photodetector, and the scanning Standa XY stage, connected via a Thorlabs optical fiber (M28L03). (b) A close-up view of the collimator directing collimated light onto the CdS photodetector. (c) The photocurrent mapping image generated using code created with ChatGPT. The inset shows an optical image of the CdS photodetector.

To demonstrate the effectiveness of our methodology, we employed ChatGPT-4.1 to develop an autonomous AI agent capable for laboratory instrumentation control. The agent itself was built using the same iterative, prompt-driven approach described in Figure 2. As shown in Figure 5a, we began by establishing a Python interface to the OpenAI API and then used ChatGPT to generate the initial version of the agent code.

The resulting agent, illustrated in Figure 5b, operates in a closed loop: it sends prompts to the OpenAI API, receives Python code in response, executes the code to interact with a Keithley 2450 SourceMeter, and then uses the resulting output or error messages to formulate refined follow-up prompts. This process continues autonomously until the task is completed.

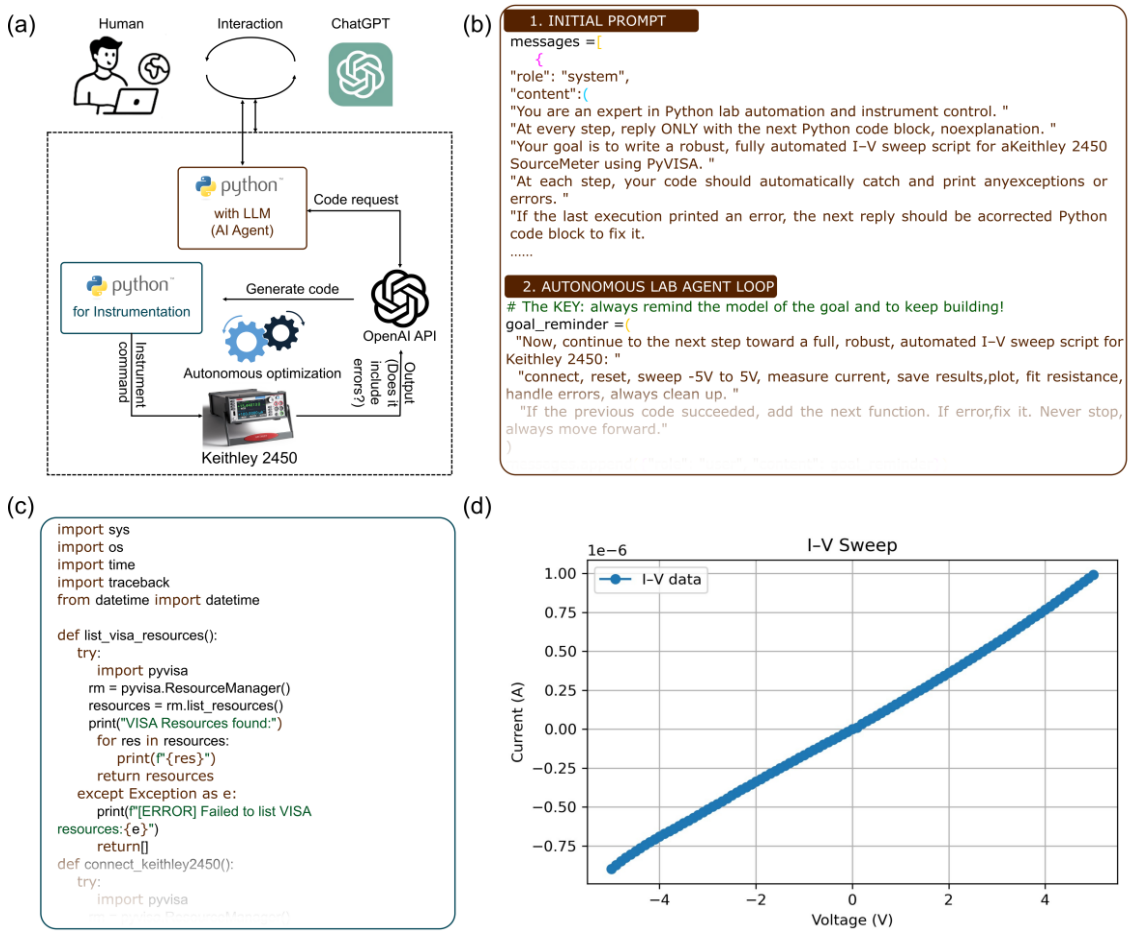

**Figure 5. Towards a fully autonomous laboratory workflow for instrumentation control using an AI Agent.** (a) Overview of the automated control pipeline: a human operator interacts with ChatGPT to generate code that interfaces with the Keithley 2450 SourceMeter via the OpenAI API. The system iteratively refines the code based on execution feedback. (b) Structure of the AI agent, including the initial system prompt and the autonomous execution loop guided by goal reminders. (c) Example Python code segment generated by the agent to identify VISA resources and initialize communication with the instrument. (d) Current–voltage (I–V) sweep measured using the final Python script produced by the agent, showing successful instrument control and data acquisition.

The agent's operation relies on a structured sequence of messages sent to the OpenAI API: a system message defines its role and behavior (e.g., *"You are an expert Python lab* *automation agent..."*), while user messages provide instructions and guide the task (e.g., *"Start by listing VISA resources..."* and *"Continue building and refining the I–V sweep script until complete"*). In this framework, the system message sets the assistant's identity and constraints, whereas user messages simulate instructions from a human operator. Through this iterative dialogue, the agent progressively generated a series of executable scripts (autolab_code_iter*.py) to automate an I–V sweep with the Keithley 2450, as shown in Figure 5c. The final result, the I–V characteristics of a photoresistor, is shown in Figure 5d. Full agent logs and generated code are provided in the Supplementary Information.

While our ChatGPT-based agent demonstrates reliable performance in automating stepwise experimental tasks such as I–V sweeps, current LLMs are not yet optimized for real-time control or experiments requiring rapid feedback and sub-second decision-making. Additionally, LLM-based automation raises ethical and safety concerns, including the risk of misoperation, data security, and regulatory compliance issues. To mitigate these risks, we recommend testing generated code in sandboxed environments under human supervision, adhering to laboratory safety protocols. For applications involving sensitive data, the use of locally deployed models is advisable. Future developments may benefit from hybrid architectures combining LLM agents with real-time control systems and rule-based safety layers.

CONCLUSIONS

This study demonstrates the transformative potential of large language models, exemplified by ChatGPT, in laboratory automation. By enabling researchers to efficiently program and control complex instrumentation, LLMs address a critical challenge in experimental science. Our case study of a single-pixel camera system and scanning photocurrent mapping system underscores the feasibility and effectiveness of this approach, paving the way for broader adoption of AI-driven tools in scientific research. The integration of LLM-based tools into laboratory workflows holds the promise of fostering innovation and accessibility, driving a new era of flexible, efficient, and customized scientific experimentation.

MATERIALS AND METHODS

*Code generation with AI language models:* ChatGPT (GPT-4o, and o3, ChatGPT 4.1) has been used to generate the code used in this manuscript.

**Supporting Information:** Detailed code examples, ChatGPT prompt links, and a recommended workflow for end-to-end measurement automation for controlling a Keithley 2450 SourceMeter and a Standa XY stage in MATLAB. Sample scripts are also provided to facilitate replication of the experimental procedures for the AI agent. The supporting data and code are available at Zenodo: https://doi.org/10.5281/zenodo.15065601

AUTHOR INFORMATION

**Corresponding Authors**

(YX) yxie@ xidian.edu.cn   (ACG) andres.castellanos@csic.es

**Author Contributions**

Y.X. and A.C.G. conceived the idea and designed the experiments. Y.X. performed the experiments. Y.X., K.H. and A.C.G. prepared the figures. Y.X. and A.C.G. analysed the data and wrote the manuscript.

**ACKNOWLEDGEMENTS**

This work was supported by the Open Fund of State Key Laboratory of Infrared Physics (Grant No. SITP-NLIST-ZD-2024-01). A.C-G. acknowledges funding by the Ministry of Science and Innovation (Spain) through the projects TED2021-132267B-I00, PID2020-115566RB-I00, PDC2023-145920-I00 and PID2023-151946OB-I00 and the European Research Council through the ERC-2024-PoC StEnSo (grant agreement 101185235) and the ERC-2024-SyG SKIN2DTRONICS (grant agreement 101167218). The author (Y.X.) acknowledges Prof. Ju Li (MIT) and Dr. Álvaro Blanco (ICMM-CSIC) for enlightening discussion.